\documentclass{article}

\usepackage[version=4]{mhchem}
\usepackage{siunitx}
\usepackage{longtable}
\usepackage{booktabs}
\usepackage{xtab}
\usepackage{tabularx}
\usepackage{todonotes}
\usepackage{graphicx}
\usepackage{subcaption}
\usepackage{lineno}
\usepackage[numbers]{natbib}
\usepackage{url}
\usepackage{hyperref}
\usepackage{tcolorbox}
\usepackage[utf8]{inputenc}

\DeclareSIUnit\Molar{M}

\title{Four Shades of Life Sciences: A Dataset for Disinformation Detection in the Life Sciences}

\author{
    Eva Seidlmayer$^{1,\ast}$ \and
    Lukas Galke$^{2}$ \and
    Konrad U. Förstner$^{1,3}$ \\
    \small{$^{1}$ZB MED -- Information Centre for Life Sciences} \\
    \small{$^{2}$University of Southern Denmark} \\
    \small{$^{3}$TH Köln -- University of Applied Sciences} \\
    \small{$\ast$Corresponding author: seidlmayer@zbmed.de}
}

\begin{document}

\maketitle
\footnotetext[1]{Present address: ZB MED Information Centre for Life Sciences, Gleuelerstrasse 60, 50931 Cologne, Germany}
\begin{abstract}
Disseminators of disinformation often seek to attract attention or evoke emotions -- typically to gain influence or generate revenue -- resulting in distinctive rhetorical patterns that can be exploited by machine learning models. In this study, we explore linguistic and rhetorical features as proxies for distinguishing disinformative texts from other health and life-science text genres, applying both large language models and classical machine learning classifiers. 
Given the limitations of existing datasets, which mainly focus on fact checking misinformation, we introduce Four Shades of Life Sciences (FSoLS): a novel, labeled corpus of 2,603 texts on 14 life-science topics, retrieved from 17 diverse sources and classified into four categories of life science publications. The source code for replicating, and updating the dataset is available on GitHub: \url{https://github.com/EvaSeidlmayer/FourShadesofLifeSciences}
\end{abstract}

\section{Introduction}
The deliberate spread of disinformation poses an increasingly significant challenge to democratic societies \citep{pariser_filter_2011, europaische_union_free_2013, vaidhyanathan_antisocial_2018, goldstein_information_2019, gensing_fakten_2020, budak_misunderstanding_2024}.
Beyond the political realm, disinformation also has a major impact on science, in particular on health and life sciences. In this paper, we adopt the Global Disinformation Index's (GDI) definition of disinformation as "deliberately misleading information, knowingly spread, or the omission of certain facts in service of a particular narrative"~\citep{global_disinformation_index_defunding_2023}. The intentional nature of the dissemination distinguishes disinformation from misinformation. 

False and misleading claims in the life sciences encompass far more than just the recent COVID-19 pandemic. 
They also include fabricated assertions about the effects of consuming tobacco in the second half of the twentieth century \citep{tan_misinformation_2020, oreskes_merchants_2012}, bogus treatments for chronic diseases such as cancer \citep{shi_false_2019}, and misleading narratives around climate change and vaccination \citep{mcgarity_bending_2008}. Such tactics stretch back centuries, from medieval conspiracies blaming minorities for poisoned wells to today's vaccine hesitancy, which the World Health Organization (WHO) now lists among the top ten global health threats\footnote{\url{https://www.who.int/news-room/spotlight/ten-threats-to-global-health-in-2019}, retrieved 2025-06-04}. The Global Disinformation Index even warns that scientific fraud undermines human rights \citep{global_dininformation_index_how_2023}.
Although disinformation predates the computer era, social networks and the internet have transformed it into one of the most pressing challenges of our time.

Recognizing and preventing disinformation can be challenging. Even rigorous quality control cannot completely filter it out, as disseminators employ sophisticated strategies to spread their narratives \citep{reed_disinformation_2021}. 
Broad suppression of dissenting or suspicious positions is problematic, as it could inadvertently affect borderline cases, emerging trends, or critical perspectives, which would be tantamount to censorship.
Empowering users to critically evaluate information is therefore key to helping them navigate the information era effectively \citep{goldstein_information_2019}. 

Automated fact-checking is a common strategy for managing misinformation, yet it poses significant challenges (see Section~\ref{section:discussion}). Although the accuracy of verifying individual statements may be improving, and fact-checking now extends to complete arguments and long-form texts, the core problem of establishing a reference corpus for truth remains.
This challenge is particularly acute in the case of scientific texts, which are often novel and controversial. Defining truth through such a reference corpus is particularly unsuitable for this kind of content. Moreover, even when false information is correctly identified, understanding disinformation still requires us to consider the intent behind its dissemination \citep{guo2022survey}. 
Since the spread of false information is often motivated by financial gain, power, or a desire to promote a product or issue \citep{RYAN20207}, it tends to employ rhetorical methods that are designed to evoke emotion or attract attention \citep{schutz-etal-2024-gerdisdetect}. Previous research has shown that the language of fake news uses more social, colloquial and temporal
words \citep{perez-rosas_automatic_2017}.  

In recent years, machine learning techniques have demonstrated significant potential for analyzing and evaluating language and textual data. Such techniques can be used to augment metadata with quality indicators to help human evaluators assess the validity of a source. However, effective use of machine learning requires meticulously curated training data \citep{bender_dangers_2021}.

This paper explores whether purely linguistic information can serve as a proxy to distinguish disinformation from other text categories. While the proposed approach should not be the sole basis for document assessment, it can provide valuable insights to support the evaluation process.



Although dissemination of disinformation and misinformation in the life sciences is a highly significant social issue, scientific texts remain underrepresented in existing research and datasets, which focus predominantly on political (dis)information. The COVID-19 pandemic and associated ''infodemic'' \citep{who_lets_2020} prompted increased interest in health and science misinformation, leading to several new datasets. However, these largely consist of social media statements, with full-length articles and other genres being relatively scarce.
Many of these datasets -- several of which have been revised (see Appendix~\ref{Appendix}, an elaborated overview provides \citep{Alghamdi2024}) -- focus on brief social-media statements rather than complete texts, target specific topics (e.g. COVID-19, tobacco, global warming), and rely on simple binary classifications (misinformation/legitimate, real/fake) that overlook the nuances of complex arguments. Moreover, many existing datasets lack the rigorous curation required for reproducible research or are otherwise inaccessible, highlighting the need for a new approach to dataset development.

\bigbreak

By incorporating full-text articles, the "Four Shades of Life Sciences" dataset fills a gap among the few disinformation datasets that exist in the field of life sciences. 
Furthermore, its classification system -- including vernacular and alternative scientific texts -- captures other deviant forms of evidence-based science, allowing for language-style and content analysis of borderline cases in downstream and data-mining tasks. As some of the data cannot be freely published, we provide a manual in our GitHub repository for independent reconstruction of the dataset (\url{https://github.com/EvaSeidlmayer/FourShadesofLifeSciences}).  
 
Four Shades of Life Science considers different degrees of information  across different genres targeting different audiences (specialists, general public, esoteric practitioners), and thereby enables downstream projects to achieve more fine-grained classification that goes beyond simplistic true-or-false decisions about information and disinformation.  

The remainder of this paper is organized as follows. First, we outline our four main contributions to advancing research in this area. Next, we describe our dataset in detail and provide a preliminary linguistic analysis of its four classes using random forest and TF-IDF methods. 
Subsequently, we present experimental results from training diverse models on the dataset, including both large language models and classical machine learning architectures. This includes analysis of the effects of previously unseen topics on prediction quality, the impact of varying text lengths processed by the models, and the use of citation frequency as a proxy for text quality. 
We conclude with a discussion of key findings and a critical evaluation of dataset preparation and cleaning protocols.

Our work makes four key contributions to advancing research into the automated detection of disinformation:

\begin{itemize}
    \item Firstly, the Four Shades of Life Sciences dataset itself (FSoLS-25-v5), which contains 2,603 texts on 14 life-science topics from 4 text classes, retrieved from 17 data sources. The code provided in the GitHub repository (\url{https://github.com/EvaSeidlmayer/FourShadesofLifeSciences}) enables dataset replication. 
    \item Secondly: BioBERT, fine-tuned on FSoLS for 3 epochs and enhanced with a sliding-window approach to process 2.5k tokens instead of the default of 512 tokens. This outperformed all other models tested (see Section \ref{section:slidingwindow}).
    \item Thirdly, the strong results achieved by a classical machine learning model, which came close to the performance of BioBERT. The best-performing model was a linear support vector classifier (see Section \ref{section:smallmodels}). Its lower computational emissions and greater explainability make it a preferable alternative to larger models. 
    \item Finally, our conclusion that using syntax and semantics as proxies is a reasonable strategy for distinguishing disinformation from other text categories.
    
\end{itemize}

\begin{table}[!htbp]
\centering
\caption{Dataset statistics by text class}
\label{tab:textstat}

\begin{tabular}{ l S[table-format=5.0] S[table-format=4.0] }
\toprule
{Text class} & {Average text length (characters)} & {Number of items} \\
\midrule
Scientific texts & 30136 & 687 \\
Vernacular texts & 5301 & 715 \\
Alternative scientific texts & 14424 & 667 \\
Disinformative texts & 8454 & 534 \\
\bottomrule
\end{tabular}
\end{table}

\begin{figure}
    \centering
    \includegraphics[width=\textwidth]{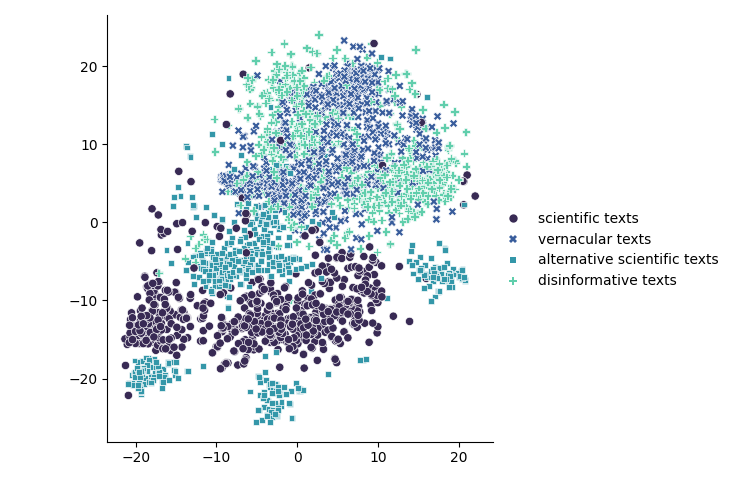}
    \caption{\label{fig:tsne}Vectorized t-SNE representation of FSoLS applying doc2vec with regard to FSoLS categories.}
   
\end{figure}

\section{Four Shades of Life Science (FSoLS) dataset}\label{ssec:num1}

The Four Shades of Life Science (FSoLS) dataset contains various genres of health-related texts with a particular focus on disinformation. Unlike previous approaches, we capture four different shades of information by classifying texts into four genres, or classes. To our knowledge, no other dataset within the life sciences offers a spectrum of text genres, including disinformation.
In contrast to other datasets, FSoLS includes full-text articles, not just short statements (see \ref{Appendix}).
FSoLS also emphasizes a balanced mix of topics, data sources, and categories. Currently, the dataset consists of 2,603 items covering 14 general topics from 17 data sources (see Tables \ref{tab:topics} and \ref{tab:amount_datasources}), and we plan to expand it further in the future. We deliberately chose general topics to ensure that the models learn to recognize text styles rather than topic-specific content. As expected, analysis of average text lengths reveals that scientific texts are the longest, averaging over 30,000 characters. These are followed by alternative scientific texts, while disinformative and vernacular texts are the shortest, each averaging just over 5,000 characters.



Figure \ref{fig:tsne} shows a t-SNE visualization \citep{tsne} of documents, embedded with doc2vec. The coloring was carried out subsequently and represents the text classes.
For the t-SNE algorithm, we used a default perplexity of $30$.
The dataset shows clear clustering corresponding to the designated classes and data sources.

\subsubsection{Class characteristics}\label{section:classcharacteristics}
To understand the language characteristics of the four classes, we performed TF-IDF analysis alongside an examination of random forest decision rules. 
This linguistic analysis of the four classes provides insights into the decision-making processes of the classification models, thereby improving machine learning explainability and interpretability.   

By comparing TF-IDF terms with terms frequently used in the entire dataset, it is possible to identify which terms are distinctive to the specific context of the dataset classes. 
To mitigate any bias resulting from the focus on specific topics, the dataset was compiled with an equal number of texts for each topic (see Table \ref{tab:topics}). This strategy ensures that the dataset's categorization relies on linguistic and semantic characteristics rather than subject matter.     

Using random forest analysis on the dataset, we identified key terms relevant to classification decisions. A total of 100 decision trees were generated and the terms from the decision rules were extracted, counted, and ranked. This allowed us to identify the most influential terms driving classification toward specific classes. In this way, we were able to systematically evaluate both the decision rules and their associated classes.

In general, terms that indicate a textual structure are highly relevant for classification (''references'': 34 times, ''introduction'': 25, ''abstract'': 24, ''discussion'': 18,  ''conclusion'': 15,  ''conclusions'': 13, ''background'': 12). 
These terms are predominantly associated with scientific or alternative scientific documents, though ''reference'' also appears frequently in disinformation texts.   
Bibliographic terms are similarly important, including ''published'': 24, ''journal'': 24, ''accepted'': 24, ''correspondence'': 21, ''review'': 21, ''corresponding'': 14, ''table'': 15.
Compared to the TF-IDF analysis, which showed the term ''et alii'' as a category indicator, the term fragments ''et'' (20) and ''al'' (22) are also relevant decision indicators. 
In addition, common linguistic elements of colloquial language identified through TF-IDF analysis -- including ''re'' (short form of ''are'') (18), ''your'' (26), ''our'' (17), and ''story'' (12) -- proved to be pertinent features in the construction of decision trees particularly in distinguishing vernacular and disinformative text styles. This is consistent with the findings of \cite{perez-rosas_automatic_2017}, who identified the language of fake news as being more socially oriented and verbally expressive compared to other types of content.


Terms such as ''herbal'' (14 times) and ''alternative'' (17) are reasonable indicators of the style of alternative scientific texts. However, the words ''homeopathic'' (22) and ''homeopathy'' (16) may be overly explicit and should be removed in subsequent versions of the dataset. All of these concepts clearly push the decision tree in the direction of alternative scientific (class 0). 
The terms ''Delhi'' (20) and ''India'' (14) reflect the large number of studies in alternative sciences originating from India. It is important to address such regional references in subsequent data-cleaning efforts. However, ''Indian'' also shows a high weight for the scientific text class.    
Additional terms that tend to indicate alternative scientific texts include ''college'' (16) and ''medicines'' (16).

Concepts that do not clearly align with a single class but remain relevant within decision trees include ''university'' (22 times) and ''guidelines'' (12), which oscillate between the classes of vernacular text style and scientific text style.
Likewise, the terms ''cells'' (16) and ''md'' (12) are associated with both the alternative scientific and scientific text classes. The term ''doctor'' (32), which is related to ''md'', pushes the random forest decision toward the disinformative and vernacular scientific text classes, likely reflecting the use of academic titles to lend authority to the content. 

Within disinformative texts, the terms ''study'' (16), ''significantly'' (16), ''medications'' (16), ''evidence'' (16) and ''researchers'' also carry considerable weight.



TF-IDF is a statistical measure used in information retrieval that combines term frequency (''frequency of occurrence'') and inverse document frequency to assess the relevance of terms within documents in a text corpus.
This measure can also be applied to corpus classes in relation to the full corpus. By calculating the weight of words within a document class relative to the entire text corpus, we can more effectively characterize and understand the distinctive features of different document classes.

The terms that characterize each category are as follows: Unsurprisingly, scientific texts contain various specialized clinical and experimental abbreviations (e. g. ''rtg4510'', ''t2d'', etc.) as well as structural elements typical of academic articles (''et alii'', mathematical formulas, etc.). This confirms observations from the random forest analysis. 
Vernacular texts are characterized by the frequent mention of author and expert names and explicit identification of these individuals as doctorate holders. In contrast to the other text classes, vernacular texts also refer to branded products rather than generic active ingredients (e. g. ''Gentesa''), as well as specific product names in general (e. g. ''Peloton'', a fitness technology). The disinformative genre is characterized by colloquial contractions (''n't'', ''re''), a pattern that was already evident in our analysis of decision rules within the random forest model. 

Alternative scientific texts frequently feature terminology from homeopathy alongside references to approaches such as ''mistletoe'' therapy and ''acupuncture''.

TF-IDF analysis of texts from databases classified by experts as disinformation disseminators reveals frequent mention of political actors (''government'') and product names (''lupron'', ''glyphosate''). 
In the context of vaccines, both the product name ''Gardasil'' (a vaccine against the human papillomavirus) and the term ''pro-vaccine'' are characteristic of the disinformative text style.

In discussions of abortion, the terms ''infanticide'' and ''pro-life'' are employed. Concerning vaccination, a generally critical stance is evident through frequent references to the ''Vaccine Adverse Event Reporting System (VAERS)''. 
Although ''plastic'' and ''microplastics'' were not explicit keywords in the article selection, these concepts are nonetheless present in the texts. The use of the term ''whistleblower'' may reflect a self-perception among disinformation authors as bearers of hidden knowledge.

As anticipated, we were able to extract characteristic concepts for the four text genres within life sciences from the TF-IDF analysis. The identified key features corroborate the results of the random forest analysis.

\section{Model benchmarking}\label{benchmarking}
As initial baselines, we consider BERT~\citep{DBLP:conf/naacl/DevlinCLT19} (version \texttt{bert-base-uncased}, 
BioBERT~\citep{leeBioBERTPretrainedBiomedical2020} (\texttt{biobert-v1.1}), SPECTER \citep{cohan_specter_2020} and Mistral 7B \footnote{\url{https://huggingface.co/mistralai/Mistral-7B-v0.3}}. 
This selection is motivated by comparative research demonstrating that encoder-only language models are preferable to decoder-only models and graph-based models for text classification~\citep{galke2023really}. According to \citep{Alghamdi2024}, there remains a relative paucity of research examining the application of recent pre-trained transformer-based models for fake news detection.

\subsubsection{Vanilla clustering by non-fine-tuned large language models }\label{section:vanilla}
To perform classification with untrained models, we embedded the texts using each model's native tokenizer. 
For contextualized embeddings, we used the penultimate layer's activations of the model, aggregated via mean pooling over the sequence dimension.
Based on the embeddings, k-means clustering was performed, applying scikit-learn, with the aim for 4 clusters. The k-means algorithm is an unsupervised method of clustering that minimizes variances within clusters. The clusters are created by assigning quantified vectors to the center of defined prototypes of the cluster. 
When classifying the dataset using neutral models applying k-means clustering, the results were poor, with F1-scores barely exceeding 40 percent (see Table \ref{tab:baseline}). BioBERT without any fine-tuning completely failed to classify the scientific text class, and the Bert-base-uncased model likewise failed on the vernacular  text class.   

\begin{table}[bt]
\caption{\label{tab:baseline}Classifying results of neutral models: F1-score per epoch}
\begin{tabular}{ l S r }
\toprule
Model &   {Accuracy}  & {F1}\\
\midrule
Bert base uncased  & 0.0941 & 0.1046 \\
BioBERT &  0.0188 & 0.0208 \\
SPECTER &  0.4539 & \textbf{0.4446} \\
Mistral 7B  &  0.0376 & 0.0382\\
\bottomrule
 \label{tab:results_neutral_Models}
\end{tabular}
\end{table}

\subsubsection{Fine-tuning of large language models}
Subsequently, we fine-tuned each model over 4 epochs with a constant learning rate of 3e-5 using the Adam optimizer~\citep{adam}. In accordance with each model's specifications, we tokenized the text using the respective tokenizer and truncated the documents to the context length limit of the BERT models, i.e., to the first 512 tokens. 
The SPECTER model exceeds the limitation of 512 tokens by default.  

We performed a fine-tuning of different transformer models (BERT-base-uncased, BioBERT, SPECTER) with a split ratio of 85\% for the training/validation set and 15\% for the test set (1,988 items in train-val set and 478 items in test set). 
The optimization employed binary cross-entropy as appropriated for multi-label classification with 4 labels. Our experiments were carried out over 4 epochs with a learning rate of 3e-5. 
Table~\ref{tab:LLM_results} shows the results. All models surpass the accuracy of 90\% after 2 fine-tuning epochs, and then level off at 96--99\%.
To address the 512-token limitation of BERT-based models, we experimented with a sliding-window mechanism (see Section \ref{section:slidingwindow}). 

Overall, BioBERT achieved the best results after 3 epochs of fine-tuning (see Table \ref{tab:LLM_results}). One reason for BioBERT's superior performance may be its training on specialized biological terminology and jargon.

In our experiments, pre-trained BioBERT attained a weighted F1-score of 98\%, the highest among all models tested, including Mistral 7B. 

\begin{table}[bt]
\caption{\label{tab:LLM_results} Classification results on FSoLS for fine-tuned models: F1-Score per epoch}
\begin{tabular}{ l *{4}{S[table-format=1.4]} }
\toprule
Model & \multicolumn{4}{c}{Epoch} \\
 & {1} & {2} & {3} & {4} \\
\midrule

Bert base uncased  & 0.9065 & 0.9701 & 0.9748 & 0.9726  \\

BioBERT & 0.9533 & 0.9752 &  \bfseries 0.9836  & 0.9348 \\


SPECTER &   0.9462 & 0.8592 & 0.9669 & 0.9762\\

Mistral 7B  &  0.6196 & 0.9102 & 0.9456 & 0.9639 \\
\bottomrule
 \label{tab:results_trainedModels}
\end{tabular}
\end{table}


\subsubsection{Handling of previously unseen topics by trained language models}\label{section:newtopics}
 To evaluate whether the model can handle previously unseen topics, a separate test set related to three unknown topics was compiled. For comparison, a standard test set covering all topics was also created. 

For the first experimental setup, three novel keywords (''climate change'', ''pandemics'', ''urine'' -- not yet included in the dataset at this stage) were selected to assess how the trained model performs on unfamiliar subjects. Aside from the topic selection, the test data were prepared using the same methodology as the training and validation sets. The test dataset comprised 511 items distributed across the four classes. Following the approach used for the training set, we ensured balance across the four classes, with 63 elements related to ''climate change'', 260 items related  to ''pandemics'' and 188 items related to ''urine''. Since the test set was balanced across all four classes, accuracy served as the main evaluation measure.

As shown in Table \ref{tab:newtopics}, the model achieved strong and comparable performance for both the standard and unknown-topic test sets, demonstrating its ability to handle new content from familiar sources. Given these results, the items related to the three previously unseen topics were subsequently integrated into the main dataset. 

\begin{table}[bt]
\centering
\caption{Comparison of classification results on the test set for the BioBERT Model (default 512 tokens, epoch 3) including known and unknown topics}
\label{tab:newtopics}
\begin{tabular}{ l S[table-format=1.2] S[table-format=1.2] }
\toprule
Class & \multicolumn{1}{c}{F1-score} & \multicolumn{1}{c}{F1-score} \\
 & \multicolumn{1}{c}{test set} & \multicolumn{1}{c}{test set including} \\
 & & \multicolumn{1}{c}{unknown topics} \\
\midrule
Scientific text style & 0.98 & 0.99 \\
Vernacular text style & 0.83 & 0.75 \\
Disinformative text style & 0.89 & 0.83 \\
Alternative scientific text style & 0.84 & 0.96 \\
\midrule
Overall  & 0.8789 & 0.8904 \\
\bottomrule
\end{tabular}
\end{table}

\subsubsection{Sliding window for full-text analysis} \label{section:slidingwindow}

Transformer-based models, such as those based on the BERT architecture, have an input sequence length limitation of 512 tokens. 
To enhance semantic analysis, we explored whether utilizing entire texts as context-dependent information could significantly improve the model's performance. For instance, we considered elements such as conclusion sections or reference lists -- characteristic features of scientific and non-scientific publications -- which can provide valuable contextual clues for classification tasks. To address the sequence length limitation, we implemented a sliding-window approach in the transformer forward function (\cite{pappagari_hierarchical_2019},\cite{wang_multi-granularity_2018}). This allows for overlapping analyses of a text, using a window size of 512 tokens and a stride of 256 tokens. The average of these overlapping windows is then taken, creating a matrix for the complete text. The mean of the multiple trained window vectors is used for the optimizer update. 

\begin{center}
\includegraphics[width=0.5\linewidth,height=5cm]{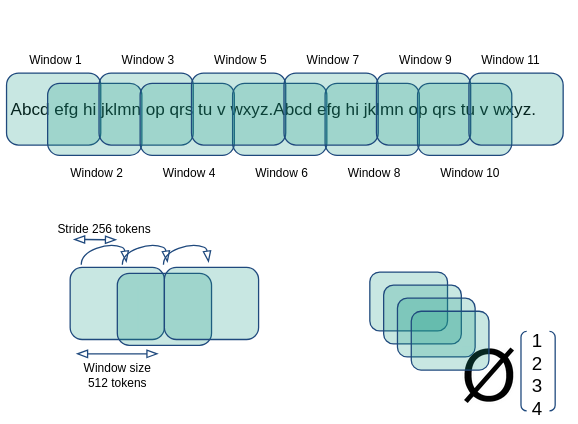}
\captionof{figure}{Sliding-window concept implemented in transformer's forward function}
\end{center}

\begin{figure}[h]\centering

\includegraphics[width=0.5\linewidth,height=5cm]{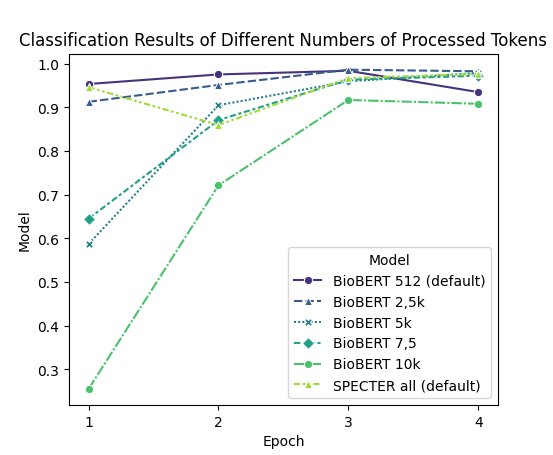}
\caption{\label{fig:slidingwindow}F1-score with different token counts for fine-tuning}
\end{figure}

BioBERT, which previously achieved the best performance with the default token limit, was further evaluated using the sliding-window approach. Token counts of 2,500, 5,000, 7,500 and 10,000 were tested (see Table~\ref{tab:slidingwindow}).   
For comparison, we also included SPECTER, which processes the tokens of full documents by default.
In line with our expectations, the modified model that takes up 2,500 tokens performed better than the basic one and also better than the SPECTER model, despite SPECTER's ability to process all tokens in a text by default (see Figure \ref{fig:slidingwindow}).   

\begin{table}[bt]
\caption{\label{tab:slidingwindow}Results with different numbers of tokens: F1-score per epoch}
\centering
\begin{tabular}{l l S[table-format=1.4] S[table-format=1.4] S[table-format=1.4] S[table-format=1.4]}
\toprule
Model & Number of tokens & \multicolumn{4}{c}{Epoch} \\
 & & 1 & 2 & 3 & 4 \\
\midrule
BioBERT & 512 (default)  & 0.9533 & 0.9752 & 0.9836 & 0.9348 \\
BioBERT & 2.5k  & 0.9125 & 0.9511 & \textbf{0.9861} & 0.9824 \\
BioBERT & 5k & 0.5868 & 0.9046 & 0.9594&  0.9786 \\
BioBERT & 7.5k & 0.6449 & 0.8705 & 0.9632 & 0.9722 \\
BioBERT & 10k & 0.2549 & 0.7212 & 0.9167 & 0.9080 \\
SPECTER & all (default) & 0.9462 &0.8592 & 0.9669 & 0.9762\\
\bottomrule
\end{tabular}
\end{table}

\subsubsection{Scientific citation frequency}\label{section:citationfreq}
The scientific text genre comprises numerous journals aggregated by PubMed Central, which generally curates high-quality, open-access full-text articles. To ensure the quality of the included articles, we experimented with citation frequency as a potential marker for scientific rigor and writing style. After collecting all PMIDs corresponding to the defined list of topics, we enriched the metadata with the respective DOIs using the NLM Entrez API \footnote{\href{https://www.ncbi.nlm.nih.gov/books/NBK25501/}{https://www.ncbi.nlm.nih.gov/books/NBK25501/, last visited April 16, 2025}}. To identify documents citing each article, we consulted the OpenCitations database \footnote{\href{https://opencitations.net/}{https://opencitations.net/, last visited June 04, 2025}}. In this way, we gathered references to topic-related articles and calculated their citation frequencies. Citation frequency was used as an indicator of scientific quality, and only articles within the top 10\% were selected for full-text retrieval and inclusion in the dataset.
To evaluate this selection strategy, we compared these highly cited texts to articles chosen without regard to citation frequency. Table~\ref{tab:citationfreqency} shows that prioritizing highly cited articles did not produce any significant difference in the training dataset's performance.  
Based on these findings, the dataset can be expanded for topics where citation counts might otherwise limit the number of articles available in other categories (see Figure \ref{fig:balanceddataset}).

\begin{table}[bt]
\caption{\label{tab:citationfreqency}BioBERT model results for scientific text style class with and without consideration of citation frequency}
\begin{tabular}{ S  S S S S S S S r}
\toprule
 {Epoch}  & {Precision} & {Recall} & {F1-Score on Scientific text style class} & {Overall F1-Score on all Classes} & \\
\midrule

\multicolumn{5}{l}{\textbf{Random Citation} Frequency (149 Support, \textbf{512} Tokens):} \\
\midrule
 1 & 1.00 & 0.99 & 1.00 & 0.9520 \\
 2  & 0.94 & 1.00 & 0.97 & 0.9789 \\ 
3 & 0.99 &  0.98 & 0.99 &  0.9810 \\ 
 4 & 0.99 & 0.97 & 0.98 & 0.9817\\
\midrule

\multicolumn{5}{l}{\textbf{Random Citation} Frequency (94 Support, \textbf{2.5k} Tokens):} \\
\midrule
 1 & 1.00 & 0.94 & 0.97 & 0.9331 \\
 2 & 0.98 & 0.99 & 0.98 & 0.9859 \\ 
3 & 0.99 &  0.99 & 0.99 &  0.9871 \\ 
 4 & 0.99 & 0.99 & 0.99 & 0.9918\\
\midrule

\multicolumn{5}{l}{\textbf{Random Citation} Frequency (101 Support, \textbf{5k} Tokens):} \\
\midrule
 1 & 0.98 & 0.88 & 0.93 & 0.7693 \\
 2  & 0.98 & 0.93 & 0.95 & 0.9393 \\ 
 3 & 1.0 &  0.87 & 0.93 &  0.9638 \\ 
4 & 1.0 & 0.96 & 0.98 & 0.9801\\
\midrule

\multicolumn{5}{l}{\textbf{Highest 10\% cited articles} (155 support, \textbf{512} tokens):} \\
\midrule
 1 & 0.98 & 0.99 & 0.99 & 0.9798 \\
 2  & 0.98 & 0.97 & 0.98 & 0.9758 \\ 
 3 & 0.96 &  0.99 & 0.97 &  0.9424 \\ 
 4 & 0.99 & 0.97 & 0.98 & 0.9707\\
\midrule

\multicolumn{5}{l}{\textbf{Highest 10\% cited articles} (97 support, \textbf{2.5k} tokens):} \\
\midrule
 1 & 0.88 & 0.99 & 0.93 & 0.8779 \\
 2 & 0.98 & 0.99 & 0.95 & 0.9652 \\ 
 3 & 0.98 &  0.99 & 0.98 &  0.9673 \\ 
 4 & 0.97 & 1.00 & 0.98 & 0.9785\\
\midrule

\multicolumn{5}{l}{\textbf{Highest 10\% cited articles} (93 support, \textbf{5k} tokens):} \\
\midrule
 1 & 0.99 & 0.96 & 0.97 & 0.7842 \\
 2  & 0.99 & 0.97 & 0.98 & 0.9527 \\ 
 3 & 0.96 &  0.99 & 0.97 & 0.9620 \\ 
 4 & 0.98 & 0.99 & 0.98 & 0.9653 \\

\bottomrule
\end{tabular}
\end{table}

\subsubsection{Classical machine learning classifiers}\label{section:smallmodels}
In addition to pre-trained large language models, we also evaluated traditional statistical classifiers based on the bag-of-words representation, which are not constrained by the 512-token limit that restricts default transformer architectures (see Table \ref{tab:bow-results}). Specifically, we tested a linear support vector machine (SVM), logistic regression, a random forest classifier, and AdaBoost classifier, all implemented in scikit-learn~\citep{scikit-learn} with default hyperparameters. 
Performance was evaluated using the F1-score and a detailed classification report. 
The dataset was preprocessed with scikit-learn's TF-IDF vectorizer (maximum 1,000 features) and with its count vectorizer. The dataset was split into 80\% training and 20\% testing sets. 
To improve the reliability of predicted probabilities, model calibration was performed using CalibratedClassifierCV with sigmoid calibration. The linear support vector classifier achieved the best results, with an F1-score of 0.97 using the TF-IDF vectorizer.   

We further analyzed the top 100 most important features for the best-performing model, the Linear SVC, across all four classes. 
Overall, these features aligned with the findings of the TF-IDF analysis and the random forest decision rules (see Section \ref{section:classcharacteristics}).
Professional terms such as ''et al'' (0.64), ''analysis'' (0.60), ''patients/patient'' (0.75/0.47), ''participants'' (0.58), ''trial'' (0.45), and structural terms like ''fig'' (0.62) are strong indicators of the scientific class.  
The vernacular class was signaled by colloquial pronouns (e. g. ''we'' (0.68), ''she'' (0.86), ''my'' (0.67)) and reporting verbs (''says'' (1.8)), reflecting the heavy use of personal statements. Interestingly, ''M.d.'' -- previously identified as a marker of legitimation -- was not among the top 100 features. 
Consistent with our earlier analyses, colloquial terms such as ''re'' (0.91) and ''your'' (0.86) were associated with the disinformative text class. Notably, ''we'', ''she'' and ''my'', which had been discussed as indicators of colloquial language, were negatively valued in the context of the disinformative text class. Additionally, the terms ''researchers'' (0.56) and ''doctors'' (0.56) appeared in this context, supporting the hypothesis that references to professionals serve as legitimation. 
The fourth class -- alternative scientific texts -- was characterized by predictable terms: ''homeopathic/homeopathy'' (2.14/0.86), ''medicine'' (1.51) and ''herbal'' (0.73). 
Interestingly, general scientific concepts like ''university'' (0.7) and  ''journal'' (0.7) were more closely associated with alternative science.

Even these classical statistical classifiers yielded good results, with a maximum F1-score of 97\% (linear support vector classifier).  
The strong performance of these smaller language models is remarkable, given their significantly lower resource requirements \citep{rigutini_performance_2024}. 

Furthermore, from an explainability perspective, we consider these statistical models preferable to large language models, which often function as black boxes. Bag-of-words-based methods provide transparent functionality, making them easier to explain to users of downstream applications. We view explainability as a crucial factor in the context of disinformation and information literacy.

\begin{table}[ht]
    \centering
    \caption{\label{tab:bow-results}F1-Score of classical machine learning classifiers}
    \begin{tabular}{l  S[table-format=1.4] S[table-format=1.4]}
\toprule
\textbf{Model} & \multicolumn{2}{c}{\textbf{F1-Score}} \\
 &  {Count Vectorizer} & {TF-IDF Vectorizer}\\
\midrule
          Linear support vector classifier (SVC)  &  0.8754 &  \textbf{0.9701}\\
          Logistic regression &  0.8347 & 0.9290 \\
          Random forest classifier &  0.9196 & 0.9403\\
          XGBoost &  0.9601 & 0.9492\\
         AdaBoost classifier &  0.7585 & 0.6591  \\
         \bottomrule
    \end{tabular}
    
\end{table}

\section{Discussion: Semantic analysis for automated genre detection in the context of disinformation}\label{section:discussion}
According to the Global Disinformation Index (GDI) definition cited above \citep{global_dininformation_index_how_2023}, disinformation is characterized by two key aspects. The information is 1) false and 2) intentionally spread. 
Most NLP research addressing disinformation has focused on fact-checking approaches that evaluate individual claims against a reference corpus to assess accuracy. However, this approach presents several challenges. The assumption that the reference corpus represents absolute truth is problematic \citep{wang_using_2023}. 
Particularly in scientific contexts, facts can be highly controversial or swiftly become outdated, yet they still remain part of ongoing scientific discussion and cannot simply be classified as ''false''. Consequently, comparison against a reference authority is unhelpful. Binary true/false statements, commonly used in fact-checking, are often inadequate in this context \citep{schutz-etal-2024-gerdisdetect}.
Furthermore, fact-checking of full texts typically involves breaking them down into individual statements, each of which is checked against the reference corpus \citep{guo2022survey}. However, problematic statements in the context of disinformation often emerge only within their broader textual context \citep{gensing_fakten_2020}. Frequently, the ''omission of certain facts'' \citep{global_dininformation_index_how_2023} transforms a text into disinformative content -- an issue that cannot be detected through examination of isolated statements. Currently, NLP can only provide meaningful support for assessing whether a statement is true or false \citep{wadden-etal-2020-fact, wang_using_2023}. 

To address these issues, our approach focuses on the second characteristic of disinformation: the intent to mislead or to spread false information. We focus on language style to detect disinformation, utilizing the semantic and syntactic characteristics of different text categories. 

Disseminators of disinformation aim to attract attention or evoke emotions, often to gain influence or for financial benefit. This results in specific rhetorical characteristics that we can use to train machine learning models. By analyzing these characteristics, our model can effectively distinguish disinformation from text genres driven by other motivations, such as the presentation of scientific findings or the communication of results to non-expert audiences. TF-IDF term analyses and investigations of random forest decision rules confirm that each genre exhibits distinct linguistic characteristics, reinforcing the validity of our approach. 

Our experimental findings support the hypothesis that training language models on genre-specific text styles is worth further exploration. 
First, we showed that untrained models do not perform adequately, making fine-tuning essential (see Section~\ref{section:vanilla}). 
Second, we found that BioBERT outperforms other encoder-only models, likely due to its pre-training on biological language. 

Further analysis revealed that incorporating larger text passages via a sliding-window approach improved BioBERT's results (see Section \ref{section:slidingwindow}). 
Processing 2,500 tokens with BioBERT after 3 epochs yielded a weighted F1-score of 0.9861. This outperformed the default models as well as other token lengths. Given that scientific texts average 28,000 characters and vernacular texts about 5,000, using even more tokens might be beneficial; this remains an area for future research.  

\begin{figure}
    \centering
    \begin{subfigure}[b]{1 \textwidth}
        \centering
        \includegraphics[width=\textwidth]{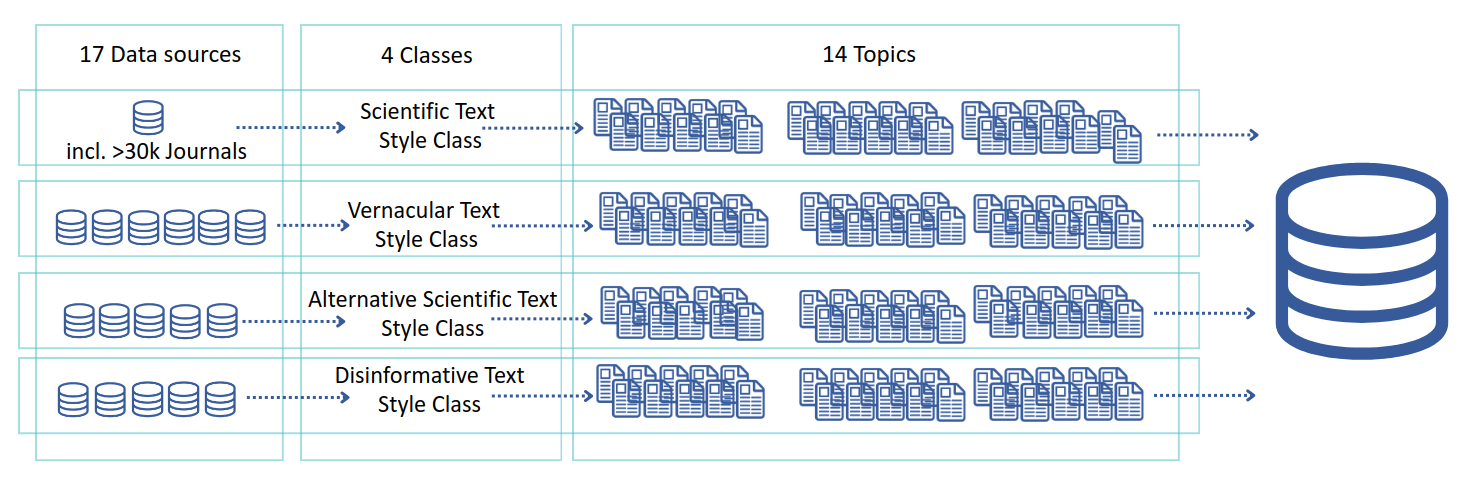}
        \caption{Dataset compilation}
    \end{subfigure}
    \hfill
    \begin{subfigure}[b]{1 \textwidth}
        \centering
        \includegraphics[width=\textwidth]{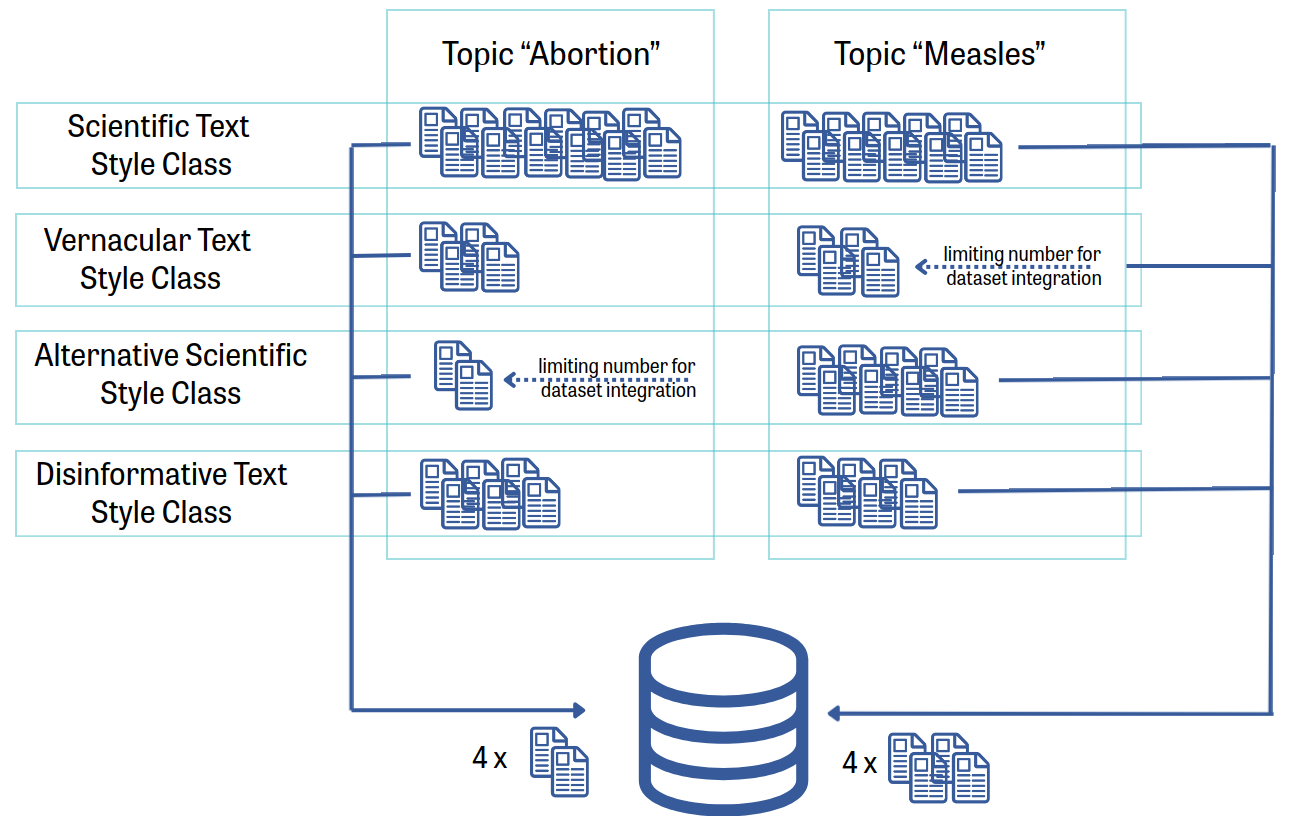}
        \caption{Balanced topics: Equal amount of text items per topic and class}
    \end{subfigure}
    \caption{\label{fig:balanceddataset}Balanced compilation of dataset with regard to (a) data sources (a) and
    (b) number of items per topic.}
    
\end{figure}

To examine the quality of texts within the scientific text genre, we conducted an experiment comparing highly cited articles with randomly selected articles (see Section \ref{section:newtopics}). The idea was to include only top-tier scientific papers identified by high citation frequency. 
However, citation frequency had no significant effect on classification performance. Consequently, we incorporated randomly chosen PubMed Central articles related to the relevant topics. This decision also allowed us to slightly expand the dataset in areas where PubMed Central had previously constrained the number of balanced items for each text genre, specifically for the topics ''turmeric'' and ''urine'' (see Figure \ref{fig:balanceddataset}).         

Balancing topics was important not only to prevent the model from learning topic-specific genre characteristics, but also to evaluate its performance on new, previously unseen topics. To this end, we created an additional, analogously compiled test set to examine how the best-performing BioBERT model (configured for 2,500-token sliding windows) handled unknown topics from known data sources. No significant differences were observed. The model's accuracy on this set was only marginally lower than on known topics, suggesting robustness; nevertheless, further evaluation of texts from entirely different sources is recommended.

Finally, classical machine learning models were also tested on their ability to classify the dataset (see Section~\ref{section:smallmodels}). All classical models performed well with both the Count Vectorizer and TF-IDF Vectorizer, with the latter generally yielding better results (except for AdaBoost). AdaBoost achieved an F1-score just above 80\%, while random forest exceeded 93\%, logistic regression surpassed 94\%, and XGBoost reached over 96\%. Although Support Vector Machine performed fairly poorly with the Count Vectorizer, it achieved the best result -- over 98\% F1 -- when paired with TF-IDF (see Table \ref{tab:bow-results}). These classical models deliver strong performance with significantly lower computational cost. Whether the roughly 2\% improvement in classification performance offered by large language models justifies their much higher resource consumption remains an open question \citep{rigutini_performance_2024}.   

An additional advantage of employing classical models is their capacity to spell out their decision-making processes. Especially in the context of disinformation, the explainability of artificial intelligence (XAI) is a critical and significant concern.
In downstream applications, it is not sufficient for a classifier to simply present the results and the designated class of a given text; it should also provide insights into the rationale behind its predictions. This level of transparency is not possible with large language models, which typically function as black boxes. In contrast, models such as the support vector machine classifier -- which achieved the best performance -- allow computation of feature importance scores and partial dependence plots, which can be presented to the users of downstream applications. This transparency enhances the interpretability of the model's predictions, allowing users to understand the factors that influenced the classification outcomes.

Although these machine learning techniques show promise for identifying disinformative content, they should not serve as the sole basis for judging an article's credibility. Classification results are statistical measurements that require further critical evaluation by users. It is therefore always important to take into account other indicators of an article's compliance with good scientific practice \citep{forschungsgemeinschaft_guidelines_2022}, such as peer-review status or citation counts. 

Finally, when classifying unfamiliar texts, our code errs on the side of caution by avoiding hard class assignments; instead, it reports similarity scores across all four classes, with the possibility of 100\% for each class.

\section{Methods and materials}

\subsection{Data sources}
The four dataset classes and the data sources used are presented in the following section.

\subsubsection{Scientific text data sources}
We define \textit{scientific} texts as those that promote evidence-based knowledge and adhere to established guidelines for good scientific practice \citep{forschungsgemeinschaft_guidelines_2022}. In the life sciences, this knowledge is backed by studies or experiments and builds on previous results. It is often characterized by sophisticated and technical language.

PubMed Central (PMC) is a repository of full-text articles curated by the US National Library of Medicine (NLM). It encompasses hundreds of thousands of scientific articles published in journals in the biomedical and life sciences fields. Most of these journals are peer reviewed and highly respected. Articles for the FSoLS dataset were retrieved for each topic using specific Medical Subject Headings (MeSH) terms via the Entrez API\footnote{\url{https://www.ncbi.nlm.nih.gov/books/NBK25501/}, retrieved June 04, 2025}.

\subsubsection{Vernacular text data sources}
\textit{Vernacular} texts make specialized knowledge accessible to broader audiences by simplifying scientific knowledge or presenting it in summary form. The language is characterized by simple sentence structures and the use of colloquial rather than technical terminology.  

Vernacular science texts were collected from MedlinePlus website, Harvard Health Publishing, and the Web MD website, as well as Men's Health and Women's Health magazines. 
MedlinePlus provides health information for patients and non-professionals. Like PMC, MedlinePlus is also maintained by the NLM. The articles on the website cover various health topics. 

Harvard Health Publishing, is a website operated by Harvard Medical School, provides patient information. Harvard Medical School is associated with the prestigious Harvard University. 

WebMD is a comprehensive online resource that provides reliable and up-to-date health information. It covers a wide range of topics, including medical conditions, medications, healthy living tips, and wellness advice. The website features articles written and reviewed by healthcare professionals, tools for symptom detection, information on drugs and supplements, and a directory of healthcare providers. WebMD is published by WebMD Health Corp, which is a subsidiary of Internet Brands.

Men's Health and Women's Health are popular magazines on health and lifestyle topics. Both magazines rank among the highest-circulation popular health magazines in the English-speaking market.

The Mayo Clinic is a US non-profit organization and operates several hospitals, mainly in the USA. The hospitals have repeatedly been awarded. The data set includes articles from the patient literature on the Mayo Clinic website from the section ''Healthy Lifestyle. Information and tools for a healthy lifestyle''.

\subsubsection{Alternative scientific text data sources}
We also include alternative science, such as \textit{homeopathy and anthroposophic} medicine, as a distinct category in the dataset. Evidence-based studies are not considered mandatory in this community. 
Derived from the Greek words hom\'{o}ios, which means like or similar, and p\'{a}thos, which means suffering, homeopathy is based on the idea that "like cures like". It employs highly diluted substances to treat a wide range of medical conditions. 

Many researchers (e.g. Quackwatch\footnote{\href{https://quackwatch.org/}{https://quackwatch.org/}}) regard homeopathy and anthroposophic medicine as inaccurate sciences because they do not adhere to conventional scientific principles. For instance, in 2021 the Congress of the German Society of Internal Medicine addressed homeopathy within a broader discussion on science denial and fake news in the medical arena ~\citep{schenk_internistenkongress_2021}.
Proponents of homeopathy argue that it has been used for centuries and that there are anecdotal success stories, but these can be explained by placebo effects. 
Alternative medical treatment itself is generally considered harmless, even if ineffective. Therefore, the mainly negative consequences of alternative medical approaches arise when it is claimed that this has a health-promoting effect, whereupon patients discontinue scientific medical treatment \citep{_honey_vermeintliche_2023}.

The ''Journal of Evidence-Based Integrative Medicine'' (JEBIM, ISSN: 2515-690X) is published four times a year by SAGE Publications in open access. Articles in JEBIM are peer reviewed. Until 1995, JEBIM was published as the Journal of Evidence-Based Complementary and Alternative Medicine \mbox{(JEBCAM)}. The publisher characterizes the journal as ''hypothesis-driven and evidence-based research in all areas of integrative medicine''.
''BMC Complementary Medicine and Therapies'' (ISSN: 2662-7671) is a peer-reviewed open-access journal published by Springer Nature which has the stated goal of publishing research articles on ''interventions and resources that complement or replace conventional therapies''\footnote{https://bmccomplementmedtherapies.biomedcentral.com/}. 
The ''International Journal of Homeopathic Sciences'' (ISSN: 2616-4493) is another peer-reviewed open-access journal that aims to ''motivate researchers in the field of homoeopathic sciences''\footnote{https://www.homoeopathicjournal.com}. The Editorial Board consists primarily of members from India. 
India is one of the largest markets for homeopathy in the world.
The ''Indian Journal of Research in Homeopathy'' (ISSN: 2320-7094)\footnote{https://www.ijrh.org/} is a peer-reviewed online journal published quarterly. It has been published by the ''Central Council For Research In Homoeopathy'' in New Delhi, India since 2007. 
''The School of Spiritual Science'' is located in Switzerland but is active worldwide ''in research, development, teaching, and the practical implementation of its research findings''\footnote{https://goetheanum.ch/en/school\#introduction}. The school is supported by the Anthroposophical Society and provides a list of ''Journal Contributions on Research in Anthroposophic Medicine'' \footnote{https://medsektion-goetheanum.org/en/research/publications/journal-contributions-on-research-in-anthroposophic-medicine-2017-2019}. Where available, publications from this list were added to the corpus.

\subsubsection{Disinformative texts data source}
Disinformation is defined as deliberately misleading information, knowingly spread, that is motivated by a purpose other than truth, such as profit or a political or religious agenda \citep{global_dininformation_index_how_2023}. 
In contrast, misinformation includes both the intentional and unintentional spread of inaccurate information, including through unchecked opinions. 
\citet{waldrop_genuine_2017} distinguish between seven types of disinformation, each with varying degrees of intent to mislead: satire, misleading content, imposter content, fabricated content, false connection, false context, and manipulated content. 

Currently, a working group of the Institute of Electrical and Electronics Engineers (IEEE) is developing a standard for the credibility of news sites\footnote{\href{https://development.standards.ieee.org/myproject-web/public/view.html\#pardetail/6318}{https://development.standards.ieee.org/myproject-web/public/view.html\#pardetail/6318}, retrieved April 18, 2025.}. Although this standard does not address scientific information sources specifically, it may provide guidance for future versions of the dataset presented in this paper. No formal standard for scientificity currently exists.

For our disinformation category, we selected websites that have been widely recognized by subject-matter experts as sources of false or misleading information.

Mercola.com and ''Dr. Mercola's Censored Library'' are highly popular alternative health news websites operated by Joseph Mercola, a physician whose professional activities have generated significant controversy. The US Food and Drug Administration (FDA) issued warning letters in 2005, 2006, 2011, and 2021 to Mercola and his websites for making false claims about products purporting to ''mitigate, prevent, treat, diagnose, or cure'' diseases (FDA 2021). Mercola.com is known for spreading misinformation during the COVID-19 pandemic and promoting claims such as autism being curable through homeopathy. The ''American Council of Science and Health'' classifies Mercola's websites as ''pure garbage'' \citep{berezow_infographic_2017}.

A broad source of disinformation content is the ''empire of NaturalNews'' \citep{institute_for_strategic_dialogue_anatomy_2020}, a complex network of US-based websites ''used to create, promote, and target health and political disinformation'' \citep{institute_for_strategic_dialogue_anatomy_2020}.  
The American Council of Science and Health also categorizes the NaturalNews website as ''pure garbage'' \citep{berezow_infographic_2017}. The renowned ''Institute for Strategic Dialogue'' (ISD)\footnote{\href{https://www.isdglobal.org/}{https://www.isdglobal.org/}} conducted an in-depth analysis of this network's many domains \citep{institute_for_strategic_dialogue_anatomy_2020}. 
Created by businessman Mike Adams, who is associated with the extremist far right anti-government militia group "Oath Keepers" \citep{institute_for_strategic_dialogue_anatomy_2020}, the network consists of at least 56 domains associated with 27 present and historic real-world entities. The website ''health.news'', from which we sourced content for our dataset, is officially run by Health News Features, LLC, though the privacy policy lists Truth Publishing International Ltd., which is located in Taichung City, Taiwan~\citep{institute_for_strategic_dialogue_anatomy_2020}.

Another cluster of disinformation websites is the ''Health Impact News'' network, classified as a ''pseudoscience website'' by the Media Bias Fact Check initiative \footnote{\href{https://mediabiasfactcheck.com/}{https://mediabiasfactcheck.com/}}. Papadogiannakis et al. found that the four Health Impact News sites are published by ''Sophia Media'', which also promotes the ''National Vaccine Information Centre'' (NVIC) and ''adfmedia.org'' websites (ADF = ''Alliance Defending Freedom'') \citep{Papadogiannakis_who-funds}. The authors documented several failed fact-checks of claims from this network. NewsGuard \footnote{\href{https://www.newsguardtech.com/}{https://www.newsguardtech.com/}}, which tracks online misinformation, identified Health Impact News as one of the most influential disseminators of COVID-19 misinformation on Facebook. The ADF has been classified as a hate group by the Southern Poverty Law Center, a US civil rights initiative.\footnote{\href{https://www.splcenter.org/fighting-hate/extremist-files/group/alliance-defending-freedom}{https://www.splcenter.org/fighting-hate/extremist-files/group/alliance-defending-freedom}} The Center defines hate groups as those ''that attack or malign an entire class of people, typically for their immutable characteristics'' \citep{southern_poverty_law_center_frequently_2022}. 

The online platform InfoWars.com, hosted by radio personality Alex Jones, was accused of propagating conspiracy theories and right-wing politics driven by ideology rather than truth \citep{van_den_bulck_lizards_2020, berezow_infographic_2017}. In 2018, Jones and his content were banned from Twitter, Facebook, YouTube, the Apple Store and other major digital platforms for violating their terms of service. In addition to fake content, the website also sold supplements, including purported COVID-19 treatments \footnote{\href{https://www.forbes.com/sites/leahrosenbaum/2020/04/09/infowars-founder-alex-jones-must-stop-selling-fake-coronavirus-silver-cures-fda-says/}{https://www.forbes.com/sites/leahrosenbaum/2020/04/09/infowars-founder-alex-jones-must-stop-selling-fake-coronavirus-silver-cures-fda-says/}}. 
Its content spanned multiple categories including Featured, Special Reports, Opinions, US News, Science \& Tech, Politics, Economy, World News, Health, and Social. 
After being ordered to pay substantial fines, \mbox{InfoWars.com} went into insolvency. In December 2024, the satirical magazine The Onion purchased the website at auction\footnote{\href{https://www.nytimes.com/2024/11/14/business/media/alex-jones-infowars-the-onion.html}{https://www.nytimes.com/2024/11/14/business/media/alex-jones-infowars-the-onion.html}}.
The FSoLS dataset only includes text from the ''Health'' section of the website prior to its acquisition by The Onion.

\subsection{Compiling a balanced dataset} 
To achieve a balanced dataset, we ensured that roughly equal numbers of items from all classes were represented. We began by randomly selecting a set of topics and then compiling texts on those topics from the respective sources. Whichever class had the lowest number of texts for a topic determined the number of texts on that topic included in the other classes. This approach prevents disproportionate representation of any single topic within a class (e.g. ''COVID-19'' in the ''disinformative text style'' category), thereby avoiding scenarios where language models might learn topical cues rather than linguistic features. 

In addition, using different data sources ensures diversity within individual classes and mitigates source-specific idiosyncrasies that cannot be fully removed by data cleaning.

\begin{table}[bt]
\caption{\label{tab:topics}Topics and amount of text items in dataset}
\begin{tabular}{ l  r}
\toprule
topic  & {total amount of items} \\
\midrule
abortion & 128 \\
climate change & 38 \\
dementia & 252 \\
heart attack & 152 \\
inflammation & 256 \\
insomnia & 128 \\
measles & 68 \\
menopause & 116 \\
pandemics & 144 \\
stroke & 288 \\
tobacco & 63 \\
turmeric & 163 \\
urine & 114 \\
vaccination & 142 \\
\bottomrule
\end{tabular}
\end{table}

\begin{table}[bt]
    \centering
    \begin{tabular}{l|l|c}
        \toprule
        category & data source & amount \\
        \midrule
        scientific & PubMed Central (PMC) & 687 \\
                  & total & 687 \\
        \midrule
        alternative & Complementary Medicine and Therapies & 309 \\
                    & Indian Journal of Research in Homeopathy & 10 \\
                    & International Journal of Homoeopathic Sciences & 138 \\
                    & Goetheaneum List & 17 \\
                    & The Journal of Evidence-Based Integrative Medicine & 193 \\
                    & total & 667 \\
        \midrule
        vernacular & Harvard Health Publishing & 116 \\
                   & Mayo Clinic & 17 \\
                   & Medline Plus & 121 \\
                   & Men's Health & 115 \\
                   & WebMD & 238 \\
                   & Women's Health & 108 \\
                   & total & 715 \\
        \midrule
        disinformative & Dr. Mercola's Censored library& 137\\ 
        & Health.News & 20 \\
               & Health Impact News  & 175 \\
               & Info Wars & 12 \\
               & Natural News & 190 \\
               & total & 534 \\
        \midrule
        complete data set & total & 2603 \\
        \bottomrule
    \end{tabular}
    \caption{Composition of dataset with regard to data sources}
    \label{tab:amount_datasources}
\end{table}

\subsection{Data cleaning}
We carefully removed extraneous fragments introduced by PDF, XML, and HTML parsing, as well as standard boilerplate phrases that do not reflect category-specific syntax: 
\begin{itemize}
    \item Additional white spaces and special characters were removed from all texts.
    \item Text retrieved from websites was stripped of article-related comments and site-specific elements such as tags or links to other websites (e.g. ''Sign up'',  ''Get the latest in health news delivered to your inbox!'', ''This site is protected by reCAPTCHA and the Google Privacy Policy and Terms of Service apply.'', ''Share this page to Facebook'', ''This site requires JavaScript to run correctly.'', ''Save Article'').
    \item  Specific recurring paragraphs, such as disclaimers, were removed (e.g. ''Disclaimer: The entire contents of this website are based upon the opinions of Dr. Mercola, unless otherwise noted [...]'').
    \item  Licensing statements were removed (e.g. ''This is an open access article under the terms of the license CC BY-NC-ND 4.0'', ''Image credit: Adobe Stock''). 
    \item Frequently cited journal identifiers such as ''British Medical Journal/BMJ'' or \newline''www.biomedcentral.com'' were removed from the bibliographies, but not from the main text where they appear as references. 
    \item PDF parsing fragments, such as ''Continued from previous page'', were deleted.
    \item Information on authors (e.g. ''About the Authors Toni Golen, MD, [...]''), was removed.
    \item DOIs were also removed.
\end{itemize}

We deliberately retained math formulas and residual ''mathusepackage'' markers as indications of mathematical content. Similarly, we opted to keep some fragments of the journal structure.

To ensure no fragments from websites or PDF parsing remained in the dataset, we calculated TF-IDF scores for category texts against the full corpus.  
We cleaned the texts until we could explain all terms that appeared in the TF-IDF analysis as characteristic of their respective categories.

\section{Data availability statement}
The full release of the dataset is not allowed due to copyright law.
German law allows text and data mining of copyright-protected content (UrHG \S 60d).\footnote{\url{https://www.gesetze-im-internet.de/urhg/__60d.html}}
On our Github account you can find instructions on how to compile the data set: https://github.com/EvaSeidlmayer/FourShadesofLifeSciences.

\section{Acknowledgements}
DFG grant number: DFG-LIS: FO 984/6-1.

\appendix
\section{Appendix: Other datasets} \label{Appendix}

The spread of life-science disinformation poses significant societal challenges, yet scientific texts remain underrepresented in text-analysis and machine learning research \citep{kotonya_explainable_2020}. Most existing studies and datasets focus on political (dis-)information, such as LIAR \citep{wang_liar_2017}, the \mbox{PolitiFact} dataset \citep{h_misra_politifact_2022}, and others.
The COVID-19 pandemic and associated ''infodemic'' \citep{who_lets_2020} prompted greater interest in misinformation and disinformation in health and science, leading to the release of several datasets. However, most of these collections only contain social-media posts; long-form texts and other genres are scarce.

Kinsora et al. \citep{kinsora_creating_2017} assembled a dataset of medical misinformation from online health forums, gathering 4.7 million user comments and responses and classifying them into two categories: "misinformation" and "non-misinformation". Verification was conducted by consulting reliable sources such as Snopes.com, MedlinePlus, CDC websites, and mayoclinic.com.

The "Health and Well Being Fake News dataset" (HWB) contains 1,000 items \citep{anoop_emotion_2020}. The authors collected 500 newspaper articles from reputable sources such as CNN, Washington Post, New York Times, and New Indian Express, classifying these as legitimate texts ("real"). For the misinformation category, they gathered 500 articles on related topics from websites known for disseminating fake news, such as BeforeItsNews, Nephef, and MadWorldNews, classifying these as "fake." Despite their offer to provide the data, we received no response even after repeated requests.

In 2020, Ambesh Shekhar published a COVID-19 misinformation dataset consisting of 7,544 items, which were collected from falsehoods detected by the CoronaVirusFacts Alliance\footnote{\href{https://www.poynter.org/coronavirusfactsalliance/}{CoronaVirusFactsAlliance}}. This dataset was published on Poynter.org and is available on Kaggle\footnote{\href{https://www.kaggle.com/datasets/ambityga/covid19misinformation}{COVID19-misinformation}}.

\citet{kotonya_explainable_2020} compiled a dataset of 11,800 statements from Reuters News, Associated Press, Health News Review, and fact-checker websites (including Politifact, FactCheck, and FullFact) for the field of public health, which they evaluated using natural language processing (NLP) models pretrained on biomedical topics. They calculated the coherence between statements in order to infer the correctness or incorrectness of a statement.

\citet{aich_telling_2022} compiled a dataset of news stories and social-media posts (tweets) on global health events~\citep{aich_telling_2022}. The collected data ranges from early-20th-century events such as smallpox and the Spanish flu to HIV/AIDS, MERS, SARS, H1N1, and Ebola, continuing through to COVID-19. The dataset is organized into three classes: (i) misinformation/unverified information; (ii) information verified by less than 4 sources; (iii) information verified by more than 4 sources. Although the categorized articles from this dataset would be compatible with our framework, the authors did not respond to our repeated requests for data access; we were therefore unable to integrate their data into our dataset.

''Industry Documents Library'' is an archive of 14 million public-health-related documents spanning tobacco, chemicals, pharmaceuticals, fossil fuels, drugs, food, and opioids. Curated by the University of California, it includes genres such as advertising, manufacturing, marketing, scientific research, political activities, and correspondence, totaling 136,538,216 pages across 24,182,266 documents. Although the raw data require substantial preprocessing, this resource is well worth further exploration.

\bibliographystyle{plainnat}

\end{document}